\documentclass[
]{ceurart}

\usepackage{listings}
\usepackage{glossaries}
\usepackage{fancyvrb}
\usepackage{pmboxdraw}
\DeclareUnicodeCharacter{2190}{\ensuremath{\leftarrow}}
\newacronym{gs}{GS}{Gold Standard}
\newacronym{eq}{EQ}{Entity–Quality}
\newacronym{llm}{LLM}{large language model}
\newacronym{nlp}{NLP}{natural language processing}
\newacronym{scp}{SCP}{Semantic CharaParser}
\newacronym{owl}{OWL}{Web Ontology Language}
\newacronym{tsv}{TSV}{tab-separated values}
\newacronym{r1}{R1}{Round 1}
\newacronym{r2}{R2}{Round 2}
\newsavebox{\treebox}

\begin{document}

\copyrightyear{2026}
\copyrightclause{Copyright for this paper by its authors.
  Use permitted under Creative Commons License Attribution 4.0
  International (CC BY 4.0).}

\conference{ICBO 2026: the International Conference on Biological and Biomedical Ontologies 2026, July 15–17, 2026, Washington DC, USA}

\title{Frontier LLM-based agents can overcome the ontology curation bottleneck for natural phenotypes}

\author[1]{James P. Balhoff}[%
orcid=0000-0002-8688-6599,
email=balhoff@renci.org,
]
\cormark[1]
\address[1]{Renaissance Computing Institute, University of North Carolina, Chapel Hill, North Carolina, USA}

\author[2]{Hilmar Lapp}[%
orcid=0000-0001-9107-0714,
email=hilmar.lapp@neuromatch.io,
]
\address[2]{Neuromatch, USA}

\cortext[1]{Corresponding author.}

\begin{abstract}
  Linking free-text phenotype descriptions to ontology terms, typically referred to as phenotype annotation, is essential for the cross-study integration of comparative morphological data. This labor intensive process has heavily relied on highly trained human experts, which makes it challenging to scale and thus a key bottleneck. Dahdul et al. (2018) established a \gls{gs} of \gls{eq} annotations across seven phylogenetic studies and used it to evaluate three human curators and the Semantic CharaParser \gls{nlp} tool with ontology-based semantic similarity metrics; they reported that machine–human consistency was significantly lower than inter-curator (human–human) consistency. Here we revisit that benchmark with five frontier hosted \glspl{llm} from Anthropic and OpenAI, each operating as an “agentic curator” within a self-contained workspace that supplies the source publication PDF, the same annotation guide used by the original human curators, the four project ontologies (UBERON, PATO, BSPO, GO), and a validation script. Evaluated against the same Gold Standard, every agent fell within the range of inter-curator variability of the three trained human biocurators of the original study; the best performing agents approached but did not reach the best performing human curator. Agents substantially outperformed Semantic CharaParser on all four metrics.
\end{abstract}


\maketitle

\section{Introduction}

The vast majority of phenotypic descriptions in the comparative biology literature—including the character matrices that underpin morphological phylogenetics—are expressed in natural language. The Phenoscape project \cite{Mabee2018-qp} pioneered the application of structured \gls{eq}-style ontology annotations \cite{Mungall2010-dp} to such datasets, enabling semantic integration with model organism phenotypes \cite{Dahdul2010-vf}. Producing these annotations at scale, however, requires biocurators, human experts deeply trained in both comparative morphology and the ontologies representing requisite domain knowledge, and is thus widely recognized as a key bottleneck.

To assess the feasibility of automating the task using machines, Dahdul et al. \cite{Dahdul2018-gt} assembled a Gold Standard of EQ annotations covering morphological characters and character states sampled from seven phylogenetic studies and compared three human curators against the \gls{scp} \gls{nlp} tool \cite{Cui2015-ji}, using ontology-based semantic similarity. Each human curator annotated in two rounds—a “Naïve” round (character and character state text only) and a “Knowledge” round (with full access to the source publication and external references). In that study, machine–human similarity was significantly lower than inter-curator similarity, a gap interpreted as a limit of the NLP system. In the eight years since, large language models capable of multi-step “agentic” behavior—reading documents, searching ontologies, calling validation tools, iterating on output—have become routinely available. We re-ran the Dahdul et al. 2018 benchmark with five such models, treating each as a Knowledge-round curator with access to the source publication, the same annotation guide given to the original human curators, the four project ontologies, and a validation script.

\section{Materials and methods}

\subsection{Gold Standard dataset}

We used the Gold Standard, characters, source publications, and reference ontologies released with Dahdul et al. 2018. The GS annotations themselves were produced by the pooled expertise of all original curators, with each character and state discussed and agreed upon collectively, and thus represent a consensus reference rather than any single curator's output. The dataset comprises 203 characters (463 states) sampled from seven phylogenetic studies in vertebrate morphology, 29 characters per publication \cite{Hill2005-le,Skutschas2012-oz,Nesbitt2011-ld,Coates2001-kl,Chakrabarty2007-yz,O-Leary2013-od,Conrad2008-je}. Terms are drawn from a merged ontology built from UBERON \cite{Haendel2014-yk}, PATO \cite{Gkoutos2018-mb}, BSPO \cite{Dahdul2014-kt}, and GO \cite{Gene_Ontology_Consortium2023-br}, with curator-added relational class expressions (e.g., part\_of some X) materialized as named \gls{owl} classes to enhance subsumption-based similarity computation. Each character state is annotated with one or more EQ statements that may use atomic terms or post-composed OWL expressions. For example, the character ``Maxillary tooth crown shape" with state ``recurved" (from Hill \cite{Hill2005-le}) is annotated with the entity expression \textit{`tooth crown' and part\_of some `maxillary tooth'} (\texttt{UBERON:0003675 and BFO:0000050 some UBERON:0011593}) and the quality `recurved' (\texttt{PATO:0002211}).

\subsection{Semantic similarity metrics}

Following Dahdul et al. we compared each (human or agentic) curator's annotations to the Gold Standard using four metrics over the merged-ontology subsumer hierarchy, computed using the ELK OWL reasoner \cite{Kazakov2013-pv}:
\begin{itemize}
    \item SimJ: Jaccard similarity of the ancestor sets of the two annotations.
    \item NIC: Normalized Information Content of the most informative common ancestor.
    \item Partial Precision (PP) and Partial Recall (PR): character-state-level asymmetric metrics that, respectively, penalize extra annotations made by the test curator and missing annotations relative to the GS, with each pair of EQs weighted by SimJ.
\end{itemize}
The different metrics are detailed further in \cite{Dahdul2018-gt}.

\subsection{Agentic curation workspace}

We constructed a self-contained workspace (Fig.~\ref{fig:workspace}) within which an agent could be launched. The ‘phenotype-eq-annotation’ skill, written in the Agent Skills format \cite{agent-skills}, encodes the procedural knowledge needed to (1) read the relevant section of the source publication, (2) decompose each character/state combination into EQ rows, (3) search the four OBO-format ontology files for appropriate Entity, Quality, and Related-Entity terms, (4) construct post-composed OWL expressions in Manchester syntax when no atomic term suffices, (5) write a 10-column \gls{tsv} row pairing each ontology term opaque identifier with its label, and (6) run the validator and iterate until the output is clean.

\texttt{validate\_annotations.scala} parses each output TSV and flags unresolvable identifiers and identifier/label mismatches (a guard against term hallucination, following DisMech \cite{dismech}), obsolete classes, unbalanced parentheses, and structural errors. The annotation guide is the same “Guide to Character Annotation” \cite{phenoscape-guide} provided to the original human curators, with example annotations translated to OWL Manchester syntax for consistency with the GS data format. Note that worked examples in the guide use anatomical structures that do not appear in the GS character set.

\begin{figure}
  \centering
  \begin{Verbatim}[frame=single,fontsize=\footnotesize]

ai-annotation/
├── AGENTS.md
├── input/
│   ├── characters/         ← per-character TSVs
│   ├── papers/             ← source publication PDFs
│   ├── ontologies/         ← UBERON, PATO, BSPO, GO
│   └── annotation_guide.md
├── output/                 ← agent annotation results
├── scripts/
│   └── validate_annotations.scala
└── .agents/
    └── skills/
        └── phenotype-eq-annotation/
            └── SKILL.md 
            
  \end{Verbatim}
  \caption{Agent workspace layout. The agent is launched with \texttt{ai-annotation/} as its working tree, with character TSVs for a single publication provided in \texttt{input/characters/}.}
  \label{fig:workspace}
\end{figure}

\subsection{Models, sessions, and rounds}

One annotation session was launched per source publication per model, with the agent prompted to annotate all 29 characters in \texttt{input/characters/} for that publication using the ‘phenotype-eq-annotation’ skill. Each agent was launched in a command-line interface harness: Claude Code for Anthropic models (Opus 4.6, Opus 4.7, Sonnet 4.6) \cite{claude-code}, Codex for OpenAI models (GPT-5.4, GPT-5.4-mini) \cite{codex}. We performed two rounds of annotation:

\begin{itemize}
    \item Round 1 (R1) (Opus 4.6, GPT-5.4) was an exploratory round in which we observed  the agents' behavior and identified weaknesses in the workspace.
    \item Round 2 (R2) (Opus 4.6, Opus 4.7, Sonnet 4.6, GPT-5.4, GPT-5.4-mini) refined the skill in the following ways: mandating reading the source publication before annotating; expanding the term-search and post-composition guidance; integrating the validator into a final correction step; and instructing the agent to perform the work without spawning sub-agents. To gauge whether the R1→R2 changes improved annotation quality, the two models present in both rounds (Opus 4.6 and GPT-5.4) were re-run on the full character set under the refined workspace; the remaining three R2 models were run only against the refined workspace. Re-running the two R1 models against the refined workspace produced modest gains on SimJ/PP/PR and a small NIC decrease. Because the first 50 characters of the GS were used during workspace refinement, the cross-model comparison reported in 3.1 is restricted to characters 51–203.
\end{itemize}

\subsection{Pipeline}

We ported the original analysis pipeline from Python 2 to Python 3 and replaced its MySQL ancestor-store with an in-memory hash. We also rebuilt the merged ontology with disjointness axioms removed, after detecting 532 unsatisfiable classes; this greatly reduced the memory required for the reasoner computation.

\section{Results}

\subsection{Agent vs. human and Semantic CharaParser performance}

\begin{figure}
  \centering
  \includegraphics[width=\linewidth-80pt]{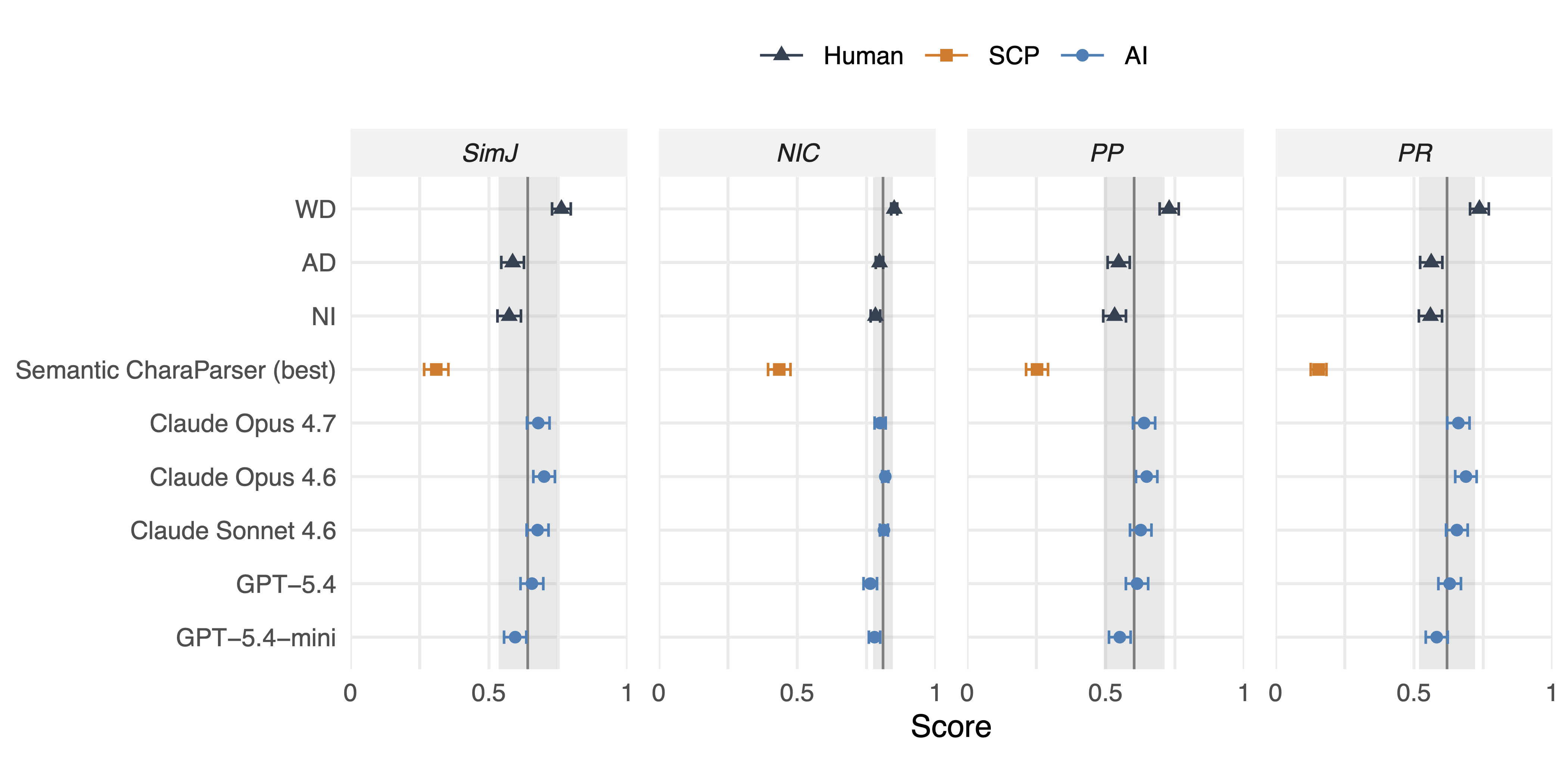}
  \caption{Mean ± 95\% CI similarity-to-GS for the three Knowledge-round human curators (WD, AD, NI), Semantic CharaParser (best variant), and the five R2 agents, on characters 51–203 (n = 344 character states). Shaded band: human curator mean ± 1 SD per metric.}
  \label{fig:main}
\end{figure}

All five agents fell within the human curator range on nearly every metric, as defined by the 95\% confidence interval being within one standard deviation of the inter-curator mean (Fig.~\ref{fig:main}). The three strongest agents—Opus 4.6, Opus 4.7, Sonnet 4.6—comfortably exceeded two human curators (AD, NI) on SimJ, PP, and PR, and matched or exceeded them on NIC. The best performing human curator (WD), who also is the most deeply trained, remained the top performer overall by a clear margin on SimJ and on PP, with the agents closer behind on NIC and PR. All five agents outperformed Semantic CharaParser on every metric, with SimJ roughly twice as high.

\subsection{Session characteristics}

Each [model, paper] session was a single agent run that produced 29 annotated character files. Recorded output-token volume—counting all generated tokens, which include internal reasoning content for both vendors—ranged from approximately 444,000 tokens for GPT-5.4 to 1.01 million tokens for GPT-5.4-mini, summed across the seven publications. Within each vendor's model family, the smaller model used more output tokens than its larger counterpart: Sonnet 4.6 generated \(1.5\times\) as many output tokens as Opus 4.6, while GPT-5.4-mini generated \(2.3\times\) as many as GPT-5.4. Output volume also varied substantially among papers, although paper-level rankings were not consistent across models.

Dahdul et al. (2015) \cite{dahdul-2015} quantified the time and effort required for phenotype annotation by human curators. They estimated a maximum rate of 13.5 characters per hour when annotation was based only on character-state text (the Naïve round of the Gold Standard study), dropping to 1.84 characters per hour when extensive literature or ontology consultation was required. Assuming a (intentionally low) labor rate of \$100 per hour, we thus estimate a cost per character between \$7.41 and \$54.35. At current (July 2026) standard Claude API list prices, the two Claude Opus models (the best performing models here) had API-equivalent inference costs of approximately \$0.26 (Opus 4.6) and \$0.43 (Opus 4.7) per annotated character. In practice, the cost was even lower, because we ran the agents within flat-rate subscriptions. Costs associated with the OpenAI agents fall within the same range.

\subsection{Locally-hosted models}

We additionally attempted the task with two open-weight models (qwen3-coder:30b, qwen3-coder-next \cite{Cao2026-ft}) served locally via Ollama \cite{ollama} on a MacBook Pro M2 Max / 64 GB. Both struggled to follow the multi-step skill—fabricating ontology identifiers, abandoning the validator loop, or terminating before completing a paper—and produced no complete annotation run, although qwen3-coder-next was significantly closer to providing usable output. We therefore exclude them from the quantitative comparison.

\section{Discussion}

Our results show that, when given the same source publication and annotation guide as a human curator, frontier general-purpose \gls{llm} agents can consistently produce EQ annotations whose semantic similarity to the Gold Standard falls within the inter-curator variability of trained human experts. The best performing agents (Claude Opus 4.6 and Opus 4.7) approach but do not yet reach the best performing human curator. This is a substantial change from the original SCP baseline for machine-based character state annotation.

The human curators in the original study worked in Phenex \cite{Balhoff2014-qd}, a purpose-built desktop application providing ontology term lookup across the same four ontologies, structured input fields that enforced the EQ schema, and semantics-based guidance—for example, flagging when a relational quality required a paired anatomical entity. Our agentic workspace provided functional analogs: the OBO files supplied term lookup, the annotation guide supplied EQ structure and modeling patterns, and the validator caught identifier and syntactic errors. The agent's role was to compose these affordances itself rather than have them pre-wired into a UI.

When comparing estimated levels of human effort required for the same kind of curation work, we find that agentic annotation using the highest performing model is not only much more time-efficient, but also very cost effective. This suggests that in the future the effort of highly trained domain experts in such a knowledgebase project can be focused at higher level tasks without sacrificing annotation quality.

One caveat for interpreting the results is that the GS dataset has been publicly available since 2018; some or all of its character states and annotations may therefore have been present in the \gls{llm}’s training data, and the ontologies themselves very likely were. However, within an annotation session one can see the AI agent looking up character terms in the ontologies, comparing labels and definitions, assessing best fit, and responding to validator feedback.

The agentic workspace described here was deliberately built as a minimal extension to the 2018 setup. If applied at production scale, the effectiveness of the workspace could be improved further by (1) a database-backed ontology query API such as OAK \cite{Mungall2026-hb} exposed as agent tools rather than as documents to skim, (2) named skills for post-composition templates (e.g., spatial qualifier, negation, magnitude-relative-to) in place of the long Markdown checklist, and (3) a logical-reasoning validation pass that classifies each composed expression with an OWL reasoner and rejects unsatisfiable constructions.

The open-weight models we were able to run locally did not yet succeed at this task. It is possible that larger open-weight models that require cutting-edge HPC-grade GPUs would succeed, and with reasonable performance, but a fair and comprehensive comparison of hosted closed-source frontier to local open-weight \glspl{llm} was beyond the scope of this study.

Finally, while the gold standard data represent a wide span of vertebrate evolution, extension of the approach to other, unrelated taxonomic groups (e.g., invertebrates, plants) remains to be established. One principal prerequisite and determinant of the quality of the result is of course the availability of a sufficiently rich anatomy ontology for the chosen taxonomic group. In the original Gold Standard study, curators were allowed to request terms be added to the respective ontologies if in their determination a concept needed for annotation was not yet present as a term. For this study we used the versions of these ontologies that included these additions, and we thus do not evaluate an agent's ability to perform this task. This may, however, be necessary for taxonomic groups where the requisite anatomy and other ontologies are less complete.

\section{Conclusion}

Eight years after the original Gold Standard was published, contemporary \gls{llm} agents—given a curation workspace that mirrors what a trained biocurator would use—perform \gls{eq} phenotype annotation of evolutionary character states at a level indistinguishable from human inter-curator variability when using semantic similarity metrics, and well above the prior automated baseline. This makes the kind of structured re-annotation of comparative morphology literature that motivated the Phenoscape project far more feasible at much greater scale, and argues for the application of agentic systems to other kinds of biocuration tasks.

\begin{acknowledgments}
We thank J. Garcia and C. Mungall for helpful feedback, as well as the other authors of the original study—W. Dahdul, P. Manda, H. Cui, A. Dececchi, N. Ibrahim, T. Vision, and P. Mabee—whose work we extend here. This work was supported by the Imageomics Institute, funded by the NSF Harnessing the Data Revolution program, award \#2118240. Code and data are available at https://github.com/phenoscape/goldstandard; the original GS dataset is archived at https://zenodo.org/records/1345307.
\end{acknowledgments}

\section*{Declaration on Generative AI}
During the preparation of this work, the authors used \textbf{Claude Opus 4.8} and \textbf{OpenAI GPT 5.5} in order to: \textbf{Drafting content}, \textbf{Paraphrase and reword}, \textbf{Formatting assistance}. After using these tools/services, the authors reviewed and edited the content as needed and take full responsibility for the publication’s content. 

\bibliography{agentic-annotation}

\end{document}